\documentclass[a4paper]{report}
\usepackage[utf8]{inputenc}
\usepackage[T1]{fontenc}
\usepackage{RJournal}
\usepackage{amsmath,amssymb,array}
\usepackage{booktabs}
\usepackage{url}

\begin{document}

\sectionhead{Contributed research article}
\volume{9}
\volnumber{1}
\year{2017}
\month{June}

\begin{article}
\title{PSF: Introduction to R Package for Pattern Sequence Based Forecasting Algorithm}
\author{by Neeraj Bokde, Gualberto Asencio-Cort\'es, Francisco Mart\'inez-\'Alvarez and Kishore Kulat}

\maketitle

\abstract{This paper introduces the R package that implements the Pattern Sequence based Forecasting (PSF) algorithm, which was developed for univariate time series forecasting. This algorithm has been successfully applied to many different fields. The PSF algorithm consists of two major parts: clustering and prediction. The clustering part includes selection of the optimum number of clusters. It labels time series data with reference to such clusters. The prediction part includes functions like optimum window size selection for specific patterns and prediction of future values with reference to past pattern sequences. The PSF package consists of various functions to implement the PSF algorithm. It also contains a function which automates all other functions to obtain optimized prediction results. The aim of this package is to promote the PSF algorithm and to ease its usage with minimum efforts. This paper describes all the functions in the PSF package with their syntax. It also provides a simple example. Finally, the usefulness of this package is discussed by comparing it to \code{auto.arima} and \code{ets}, well-known time series forecasting functions available on CRAN repository.

This paper is published in The R Journal. Kindly cite this paper as:\\

\textbf{@article\{RJ-2017-021,}

  \textbf{author} = {Neeraj Bokde and Gualberto Asencio-Cortés and Francisco Martínez-Álvarez and Kishore Kulat},
  
  \textbf{title} = {{PSF: Introduction to R Package for Pattern Sequence Based Forecasting Algorithm}},
  
  \textbf{year} = {2017},
  
  \textbf{journal} = {{The R Journal}},
  
  \textbf{url} = {https://journal.r-project.org/archive/2017/RJ-2017-021/index.html},
  
  \textbf{pages} = {324--333},
  
  \textbf{volume} = {9},
  
  \textbf{number} = {1}
\}
}

\section{Introduction}

PSF stands for Pattern Sequence Forecasting algorithm. PSF is a successful forecasting technique based on the assumption that there exist pattern sequences in the target time series data. For the first time, it was proposed in \citet{martinez2008lbf}, and an improved version was discussed in \citet{martinez2011energy}.

\citet{martinez2011energy} improved the label based forecasting (LBF) algorithm proposed in \citet{martinez2008lbf} to forecast the electricity price and compared it to other available forecasting algorithms such as ANN \citep{catalao2007short}, ARIMA \citep{conejo2005day}, mixed models \citep{garcia2007mixed} and WNN \citep{lora2007electricity}. These comparisons concluded that the PSF algorithm is able to outperform all these forecasting algorithms, at least in the electricity price/demand context.

Many authors have proposed improvements for PSF. In particular, \citet{jin2014improved} highlighted the PSF algorithm limitations and suggested minute modification to minimize the computation delay. They suggested that, instead of using multiple indexes, a single index could make computation simpler and they used the Davies Bouldin index to obtain the optimum number of clusters.

\citet{majidpour2014modified} compared PSF to kNN and ARIMA, and observed that PSF can be used for electric vehicle charging energy consumption. It also proposed three modifications in the existing PSF algorithm. First, instead of taking average of all the matched template, it only uses the most recent matched template. Second, if no match was found in the training data (which is possible), MPSF outputs the cluster center of the largest cluster as output. Third, instead of finding the optimum number of clusters starting at two, it starts k from 10\% of the total number of samples to avoid degenerate clusters.

\citet{shen2013ensemble} proposed an ensemble of PSF and five variants of the same algorithm, showing better joint performance when applied to electricity-related time series data.

\citet{koprinska2013combining} attempted to propose a new algorithm for electricity demand forecast, which is a combination of PSF and Neural networks (NN) algorithms. The results concluded that PSF-NN is performing better than original PSF algorithm.

\citet{fujimoto2012pattern} modified the clustering method in PSF algorithm. It used a cluster method based on non-negative tensor factorization instead of k-means technique and forecasted energy demand using photovoltaic energy records.

\citet{martinez2011outliers} also modified the original PSF algorithm to be used for outlier occurrence forecasting in time series. The metaheuristic searches for motifs or pattern sequences preceding certain data, marked as anomalous in the training set. Then, outlying and regular data are separately processed. Once again, this version was shown to be useful in electricity prices and demand.

The analysis of these works highlights the fact that the PSF algorithm can be applied to many different fields for time series forecasting, outperforming many existing methods. Since the PSF algorithm consists of many dependent functions and the authors did not originally publish their code, the package here described aims at making PSF handy and with minimum efforts for coding.

The rest of the paper is structured as follows. Section 2 provides an overview of the PSF algorithm. Section 3 explains how the package in R has been developed. Illustrative examples are described in Section 4. Finally, the conclusions drawn from this work are summarized in Section 5.

\section{Brief description of PSF}\label{sec:PSF}
\begin{figure}[htbp]
  \centering
  \includegraphics[width=12cm]{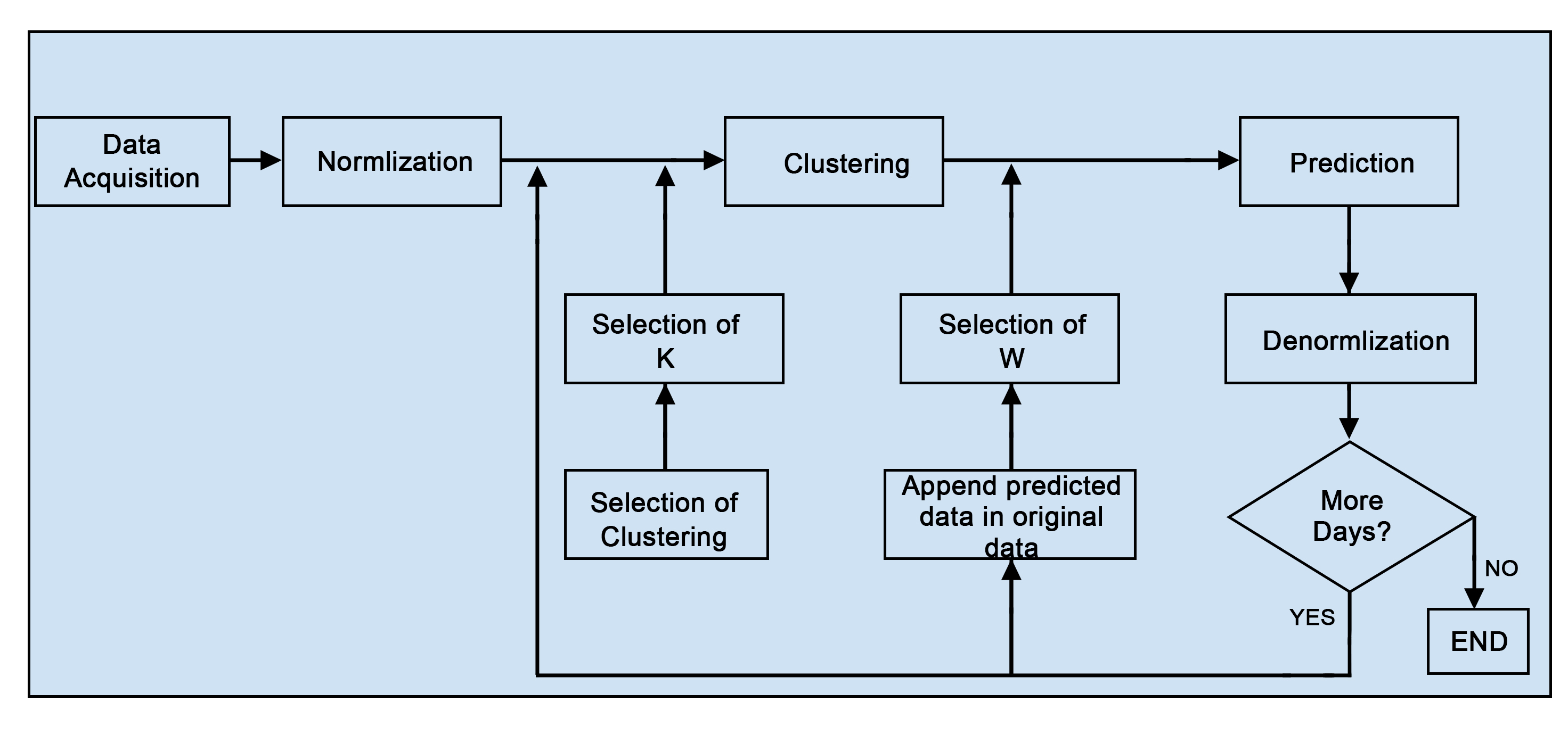}
  \caption{Block diagram of PSF algorithm methodology.}
  \label{figure:PSF}
\end{figure}

The PSF algorithm consists of many processes, which can be divided, broadly, in two steps. The first step is clustering of data and the second step is forecasting based on clustered data in earlier step. The block diagram of PSF algorithm shown in Figure \ref{figure:PSF} was proposed by \citet{martinez2008lbf}. The PSF algorithm is a closed loop process, hence it adds an advantage since it can attempt to predict the future values up to long duration by appending earlier forecasted value to existing original time series data. The block diagram for PSF algorithm shows the close loop feedback characteristics of the algorithm. Although there are various strategies for multiple-step ahead forecasting as described by \citet{Bontempi2013ahead}, this strategy turned out to be particularly suitable for electricity prices and demand forecasting, as the original work discussed.

Another interesting feature lies in its ability to simultaneously forecast multiple values, that is, it deals with arbitrary lengths for the horizon of prediction. However, it must be noted that this algorithm is particularly developed to forecast time series exhibiting some patterns in the historical data, such as weather, electricity load or solar radiation, just to mention few examples of usage in reviewed literature. The application of PSF to time series without such kind of inherent patterns might lead to the generation of not particularly competitive results.

The clustering part consists of various tasks, including data normalization, optimum number of clusters selection and application of k-means clustering itself. The ultimate goal of this step is to discover clusters of time series data and label them accordingly.

Normalization is one of the essential processes in any time series data processing technique. Normalization is used to scale data. The algorithm \citep{martinez2011energy} used the following transformation to normalize the data.

\begin{equation}
{X_j} = \frac{_{X_j}}{\frac{1}{N}\sum_{i=1}^{N} Xi}
\end{equation}

where $X_j$ is the input time series data and $N$ is the total length of the time series.

However, the original PSF original algorithm used $N=24$ hours instead of N as the total length of the time series. This presents the problem of knowing all hours for each day to assess the mean. For such reason, the original formula has been replaced in this implementation by the standard feature scaling formula (also called unity-based normalization), which bring all values into the range $[0, 1]$ \citet{dodge2003scaling}:

\begin{equation}
X_j' = \frac{ X_j - min(X_i)}{max(X_i) - min(X_i)}
\end{equation}

where $X_j'$ denotes the normalized value for $X_j$, and $i=1, \ldots, N$.

The reference articles \citep{martinez2008lbf} and \citep{martinez2011energy} used the k-means clustering technique to assign labels to sets of consecutive values. In the original manuscript, since daily energy was analyzed, clustering was applied to every 24 consecutive values, that is, to every day. The advantage of k-means is its simplicity, but it requires the number of clusters as input. In \citet{martinez2008lbf}, the Silhouette index was used to decide the optimum numbers of clusters, whereas in the improved version \citep{martinez2011energy}, three different indexes were used, which include the Silhouette index \citep{kaufman2009finding}, the Dunn index \citep{dunn1974well} and the Davies Bouldin index \citep{davies1979cluster}. However, the number of groups suggested by each of these indexes in not necessarily the same. Hence, it was suggested to select the optimum number of clusters by combining more than one index, thus proposing a majority voting system.

As output of the clustering process, the original time series data is converted into a series of labels, which is used as input in the prediction block of the second phase of the PSF algorithm. The prediction technique consists of window size selection, searching for pattern sequences and estimation processes.

Let $x(t)$ be the vector of time series data such that $x(t) = [x_1(t), x_2(t), \ldots, x_N(t)]$. After clustering and labeling, the vector converted to $y(t) = [L_1, L_2, \ldots, L_N]$, where $L_i$ are labels identifying the cluster centers to which data in vector $x(t)$ belongs to. Note that every $x_i(t)$ can be of arbitrary length and must be adjusted to the pattern sequence existing in every time series. For instance, in the original work, $x(i)$ was composed of 24 values, representing daily patterns.

Then the searching process includes the last \code{W} labels from $y(t)$ and it searches for these labels in $y(t)$. If this sequence of last \code{W} labels is not found in $y(t)$, then the search process is repeated for last \code{(W-1)} labels. In PSF, the length of this label sequence is named as \code{window size}. Therefore, the window size can vary from \code{W} to 1, although it is not usual that this event occurs. The selection of the optimum window size is very critical and important to make accurate predictions. The optimum window size selection is done in such a way that the forecasting error is minimized during the training process. Mathematically, the error function to be minimized is:

\begin{equation}
\sum_{t\epsilon TS}\left \| \overline{X}(t) - X(t) \right \|
\end{equation}

where $\overline{X}(t)$ are predicted values and $X(t)$ are original values of time series data. In practice, the window size selection is done with cross validation. All possible window sizes are tested on sample data and corresponding prediction errors are compared. The window size with minimum error considered as the optimum window size for prediction.

\begin{figure}[htbp]
  \centering
  \includegraphics[width=12cm]{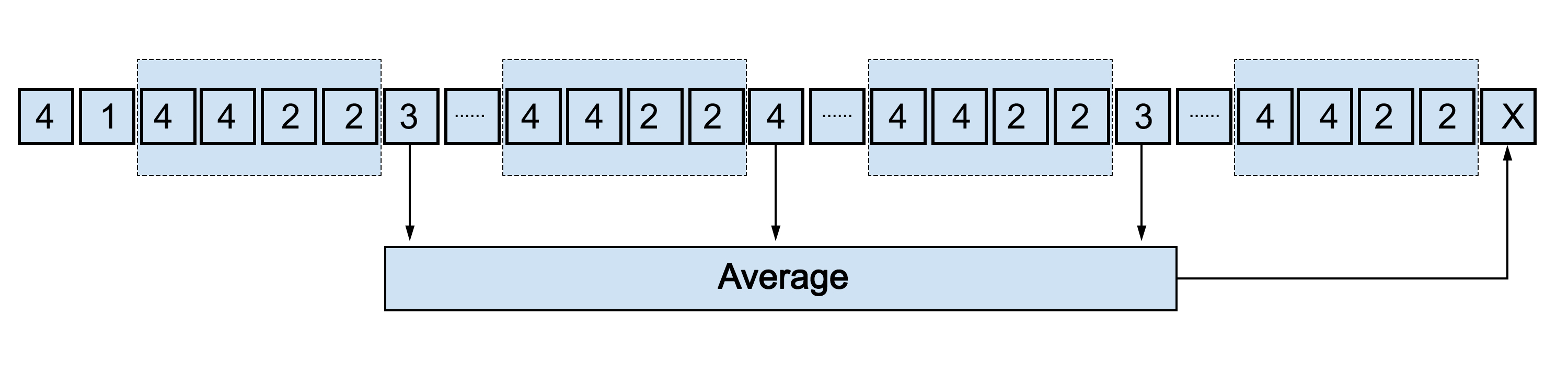}
  \caption{Prediction with PSF algorithm.}
  \label{figure:average}
\end{figure}

Once the optimum window size is obtained, the pattern sequence available in the window is searched for in $y(t)$ and the label present just after each discovered sequence is noted in a new vector, called \code{ES}. Finally, the future time series value is predicted by averaging the values in vector \code{ES} as given below:
\begin{equation}
\overline{X}(t) = \frac{1}{size(ES)} \times \sum_{j=1}^{size(ES)} ES(j)
\end{equation}

where, \code{size(ES)} is the length of vector \code{ES}.
The procedure of prediction in the PSF algorithm is described in Figure \ref{figure:average}. Note that the average is calculated with real values and not with labels.

The algorithm can predict the future value for next interval of the time series. But, to make it applicable for long term prediction, the predicted short term values will get linked with original data and the whole procedure will be carried out till the desired length of time series prediction is obtained.

\section{PSF package description}\label{sec:Package}

This section introduces the PSF package in R language. Source code along with its reference manual can be found at \url{https://cran.r-project.org/package=PSF}. With reference to dependencies, this package imports package \pkg{cluster} \citep{maechler2015package} and suggests packages \pkg{knitr} and \pkg{rmarkdown}. The package also needs the \pkg{data.table} package to improve data wrangling. PSF package consists of various functions. These functions are designed such that these can replace the block diagrams of methodology used in PSF. The block diagram in Figure 3 represents the methodology mentioned in Figure 1 with the replacement of equivalent functions used in the proposed package.

The various tasks including the optimum cluster size, window size selection, pattern searching and prediction processes are performed in the package with different functions including \code{psf()}, \code{predict.psf()}, \code{plot.psf()}, \code{optimum\_k()}, \code{optimum\_w()}, \code{psf\_predict()} and \code{convert\_datatype()}. All functions in the PSF package version 0.4 onwards are made private except for \code{psf()}, \code{predict.psf()} and \code{plot.psf()} functions, so that users could not use it directly. Moreover, if users need to change or modify clustering techniques or procedures in private functions, the code is available at GitHub (\url{https://github.com/neerajdhanraj/PSF}).

\begin{figure}[htbp]
  \centering
  \includegraphics[width=12cm]{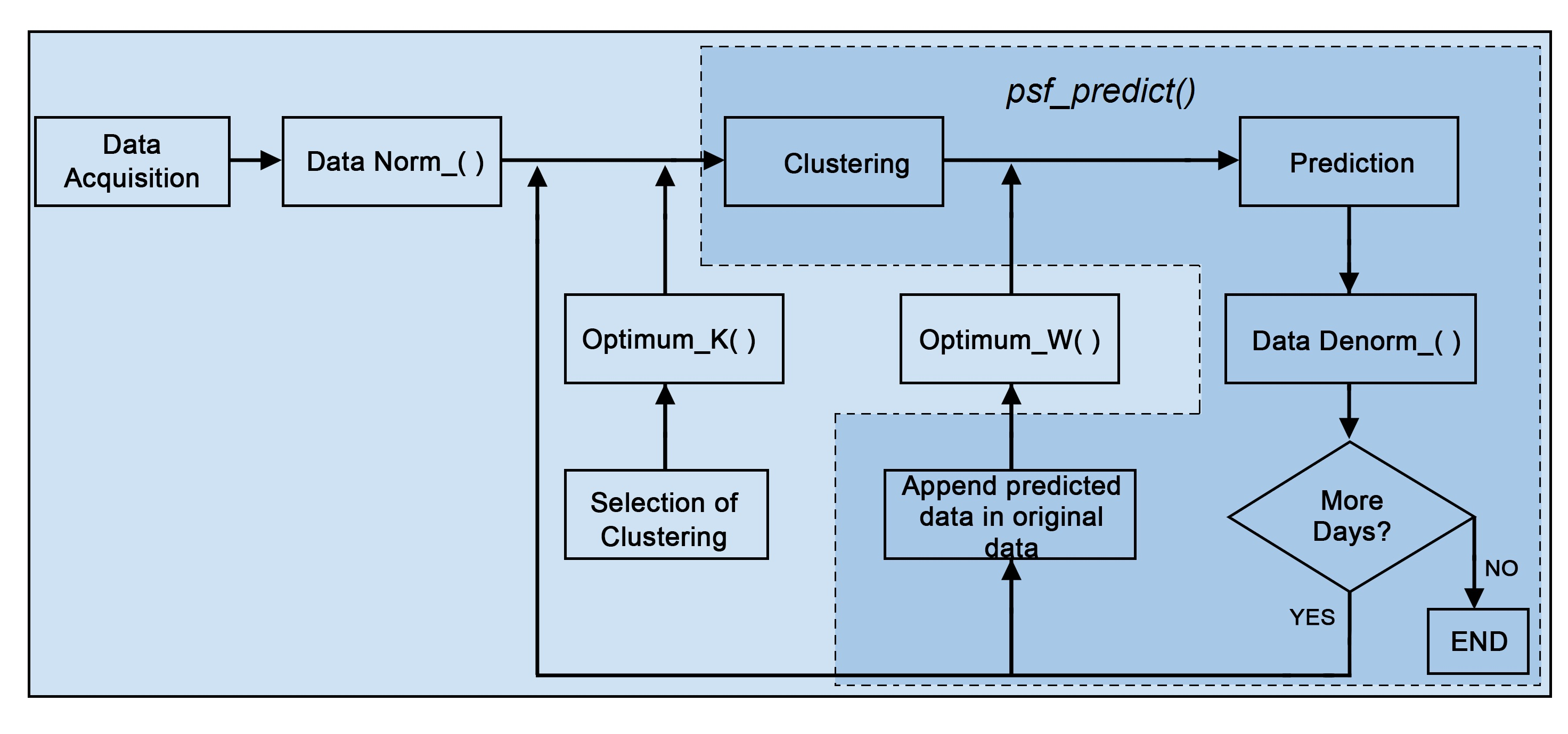}
  \caption{Block diagram with PSF package functions replacement for PSF algorithm.}
  \label{figure:rlogo}
\end{figure}

This section discusses each of these functions and their functionalities along with simple examples. All functions in the PSF package are loaded with:

\begin{example}
library(PSF)
\end{example}

\subsection{Optimum cluster size selection}
As aforementioned, clustering of data is one of the initial phases in PSF. The reference articles \citep{martinez2008lbf} and \citep{martinez2011energy} chose the k-means clustering technique for generating data clusters according to the time series data properties. The limitation of k-means clustering technique is that the adequate number of clusters must be provided by users. Hence, to avoid such situations, the proposed package contains a function \code{optimum\_k()} which calculates the optimum value for the number of clusters, (\code{k}), according to the Silhouette index. This function generates the optimum cluster size as output. In \citet{martinez2011energy}, multiple indexes (Silhouette index, Dunn index and the Davies--Bouldin index) were considered to determine the optimum number of clusters. For the sake of simplicity and to save the calculation time, only the Silhouette index is considered in this package, as suggested in \citet{martinez2008lbf}.

This \code{optimum\_k()} is a private function which is not directly accessible by users. This function takes \code{data} as input time series, in any format be it matrix, data frame, list, time series or vector. Note that the input data type must be strictly numeric. This function returns the optimum value of number of clusters (\code{k}) in numeric format.

\subsection{Optimum window size selection}
Once data clustering is performed, the optimum window size needs to be determined. This is an important but tedious and time consuming process, if it is manually done. In \citet{martinez2008lbf} and \citet{martinez2011energy}, the selection of optimum window size is done through cross--validation, in which data is partitioned into two subsets. One subset is used for analysis and the other one for validating the analysis. Since the window size will always be dependent on the pattern of the experimental data, it is necessary to determine the optimum window size for every time series data.

In the PSF package, the optimum window size selection is done with the function \code{optimum\_w()}, which takes as input the time series data, the previous estimated $k$ value, a set candidate $w$ values to search in and the cycle of the input time series. This function estimates the optimum value for the window size such that the error between predicted and actual values is minimum. Internally, this function divides the input time series data into two sub-parts. One of them is the last \code{cycle} data values which will be taken as reference to compare to the predicted values and to calculate RMSE values. The other sub-part is the remaining data set which is used as training part of the data set. The predicted values for the window size with minimum RMSE value is then taken as optimum window size. If more than one window size values are obtained with the same RMSE values, the maximum window size is preferred by the function \code{optimum\_w()}. Like the \code{optimum\_k()} function, the \code{optimum\_w()} function is also a private function and users cannot directly access it.

\subsection{Prediction with PSF}

The PSF package exposes three functions to the user. The first one, \code{psf()}, can build a PSF model from an univariate time series. The value returned by such function is a S3 object of class \code{'psf'}, which contains both original and normalized input time series along with other internal model adjustements. Once the PSF model is trained and returned, the user can invoke the S3 method \code{predict()} over the the \code{'psf'} object specifying the desired number of forecasted values via the \code{n.ahead} parameter. This method returns a numeric vector. Finally, the third function exposed is the S3 method \code{plot()}, that produces a plot including both actual and predicted values from a PSF model and a numeric vector of predictions.

Internally, the prediction procedure is composed of data processing, optimum window size and cluster size selection. The prediction is done with the private function \code{psf\_predict()}, which takes time series data, window size (w), cluster size (\code{k}) and an integer \code{n.ahead}. The value of \code{n.ahead} indicates the count up to what extend the forecasting is to be done by function \code{psf\_predict()}. The time series data taken as input can be in any format supported by R language, be it time series, vector, matrix, list or data frame. The PSF package automatically converts it in suitable format using the private function \code{convert\_datatype()} and proceeds further. All other input parameters \code{w}, \code{k} and \code{n.ahead} must be in integer format.

The function \code{psf\_predict()} initiates the process of data normalization. Then, it selects the last \code{w} labels from the input time series data and searches for that number sequence in the entire data set. Additionally, it captures the very next integer value after each sequence and calculates the average of these values. The obtained averaged value along with denormalization is considered as raw predicted value. If the input parameter \code{n.ahead} is greater than unity, then the predicted value is appended to the original input data, and the procedure is repeated \code{n.ahead} times. Finally, the processed data is denormalized and returns a time series which replaces labels by the predicted values.

The \code{psf\_predict()} function is also one of the private functions which takes input the dataset in data.table format, as well as another integer inputs including number of clusters (\code{k}), window size (\code{w}), horizon of prediction (\code{n.ahead}), and the \code{cycle} parameter, which discovers the cycle pattern followed by dataset. This function considers all input parameters and searches for the desired pattern in training data. While performing the searching process, if the desired pattern is obtained once or more times, it calculates the average of very next values for each repeated desired patterns in the whole dataset. Finally, this averaged value is considered as next predicted value and is appended to input time series data.

The \code{psf()} function uses \code{optimum\_k()} and \code{optimum\_w()} functions. The latter, in turn, uses the described \code{psf\_predict()}. The function \code{psf()} returns the PSF trained model (S3 object of class \code{'psf'}) whose contents are described later in this section. The syntax for \code{psf()} function is shown below:

\begin{example}
psf(data, k = seq(2, 10), w = seq(1, 10), cycle = 24)
\end{example}


Within the indicated syntax, the parameter \code{data} is an univariate time series in any format, e.g. time series, vector, matrix, list or data frame. Also, \code{data} must be strictly provided in numeric format.  The parameter \code{k} is the number of clusters, whereas parameter \code{w} is the window size. Finally, the \code{cycle} parameter is the number of values that confirms a cycle in the time series. Usual values for the \code{cycle} parameter can be 24 hours per day, 12 months per year or so on and it is used only when input data is not in the time series format. If input data is given in time series format, the cycle is automatically determined by its internal frequency attribute.

If the user provides a single value for either \code{k} or \code{w}, then this value is used in the PSF algorithm and the search for their optimum is skipped. Furthermore, if the user provides a vector of values for either parameter \code{k} or \code{w}, then the search for optimum is limited to these values.

This \code{psf()} function returns a S3 object of class \code{'psf'} which includes the 7 elements described below:

\begin{description}
\item[original\_series] Original time series stored to be used internally to build further plots.
\item[train\_data] Adapted and normalized internal time series used to train the PSF model.
\item[k] The number of clusters optimized and used to train the model.
\item[w] The window size optimized and used to train the model.
\item[cycle] Determined cycle for the input time series.
\item[dmin] Minimum value of the input time series (used to denormalize internally in further predictions).
\item[dmax] Maximum value of the input time series (used to denormalize internally in further predictions).
\end{description}

The \code{psf()} function initiates the conversion of any type of data format into \emph{data.table} which is an internal format for this function (this conversion is carried out by the private function \code{convert\_datatype}). Secondly, it is checked whether the input dataset is multiple of \code{cycle} parameter or not. If not, it generates a warning to state that the dataset pattern is not suitable for further study. But if the dataset is multiple of the \code{cycle} value, the normalization of the dataset is carried out and subsequently reshaped according to the cycle of time series.

This reshaped dataset is then provided to \code{psf\_predict()} function with other parameters including cluster size (\code{k}), window size (\code{w}), \code{n.ahead} and \code{cycle}, as mentioned before. In reply to this, the \code{psf\_predict()} function returns a vector of future predicted values. Since the input dataset had been previously normalized, it is required to denormalize the predicted values. This is carried out by the S3 method \code{predict.psf()}. This function returns a numeric vector with the denormalized predicted values.

\subsubsection{Plot function for PSF}

The \code{plot.psf()} function allows users to plot both the actual values of a series and predicted values obtained by a PSF model. This function takes the trained PSF model, denoted by the first parameter named \code{x}, which includes internally the original time series, obtained by \code{psf()}, along with the predicted values obtained through \code{predict.psf()}, and optionally other plot describing variables, like plot title, legends, etc. This function generates the plot showing original time series data and predicted data with significant color changes. The plot obtained by \code{plot.psf()} possesses dynamic margin size such that it can include the input data set and all predicted values as per requirements. The syntax of \code{plot.psf()} function is as shown below. The usage of \code{plot.psf()} function is further discussed in the next section.

\begin{example}	
plot.psf(x, predictions, cycle = 24, ...)
\end{example}	

\section{Example}\label{sec:Example}
This section presents the examples to introduce the use of the PSF package and to compare it with \code{auto.arima()} and \code{ets()} functions, which are well accepted functions in the R community working over time series forecasting techniques. The data used in this example are \code{nottem} and \code{sunspots} which are standard time series datasets available in R. The \code{nottem} dataset is the average air temperatures at Nottingham Castle in degrees Fahrenheit, collected for 20 years, on monthly basis. Similarly, \code{sunspots} dataset is mean relative sunspot numbers from 1749 to 1983, measured on monthly basis.

For both datasets, all the recorded values except for the final year are considered as training data, and the last year is used for testing purposes. The predicted values for final year for both datasets are discussed in this section. 

In the following examples, model training, forecasting and plotting were shown for the dataset \code{nottem}, but the procedure is the same for the dataset \code{sunspots}. In first place, the model must be trained using the PSF package. as is shown below.

\begin{Schunk}
\begin{Sinput}
library(PSF)
nottem_model <- psf(nottem)
nottem_model
\end{Sinput}
\begin{Soutput}
## $original_series
##       Jan  Feb  Mar  Apr  May  Jun  Jul  Aug  Sep  Oct  Nov  Dec
## 1920 40.6 40.8 44.4 46.7 54.1 58.5 57.7 56.4 54.3 50.5 42.9 39.8
## 1921 44.2 39.8 45.1 47.0 54.1 58.7 66.3 59.9 57.0 54.2 39.7 42.8
## ...
##
## $train_data
##             V1        V2        V3        V4        V5        V6        V7
##  1: 0.26420455 0.2698864 0.3721591 0.4375000 0.6477273 0.7727273 0.7500000
##  2: 0.36647727 0.2414773 0.3920455 0.4460227 0.6477273 0.7784091 0.9943182
## ...
##
## $k
## [1] 2
## 
## $w
## [1] 1
## 
## $cycle
## [1] 12
## 
## $dmin
## [1] 31.3
## 
## $dmax
## [1] 66.5
## 
## attr(,"class")
## [1] "psf"
\end{Soutput}
\end{Schunk}

Once the model is trained, forecasted values for the time series can be obtained using the S3 method \code{predict()} function, as is shown below.

\begin{Schunk}
\begin{Sinput}
nottem_preds <- predict(nottem_model, n.ahead = 12)
nottem_preds
\end{Sinput}
\begin{Soutput}
##  [1] 38.97692 38.71538 42.49231 46.32308 52.91538 57.97692 61.87692
##  [8] 60.19231 57.03846 49.42308 43.23846 40.21538
\end{Soutput}
\end{Schunk}

To represent the prediction performance in plot format, the S3 method \code{plot()} can be used, as shown in the following code.

\begin{Schunk}
\begin{Sinput}
plot(nottem_model, nottem_preds)
\end{Sinput}
\end{Schunk}

\begin{figure}[htbp]
  \centering
  \includegraphics[width=12cm]{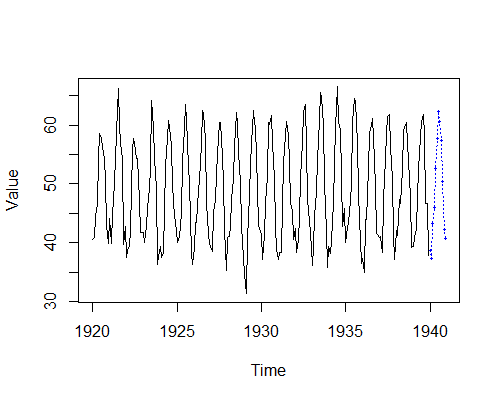}
  \caption{Plot showing forecasted results with function \code{plot.psf()} for \code{nottem} dataset.}
  \label{figure:rlogo4}
\end{figure}

\begin{figure}[htbp]
  \centering
  \includegraphics[width=10cm]{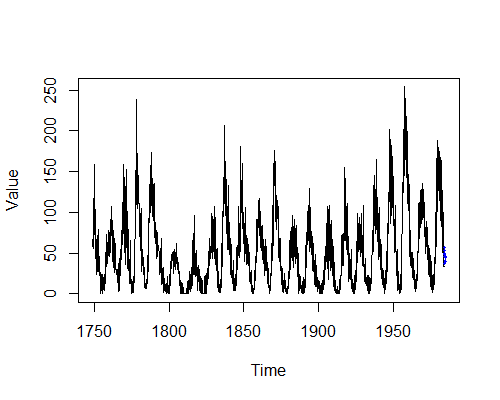}
  \caption{Plot showing forecasted results with function \code{plot.psf()} for \code{sunspots} dataset.}
  \label{figure:rlogo5}
\end{figure}

Figures \ref{figure:rlogo4} and \ref{figure:rlogo5} show the prediction with the PSF algorithm for \code{nottem} and \code{sunspots} datasets, respectively. Such results are compared to those of \code{auto.arima()} and \code{ets()} functions from \pkg{forecast} R package. \code{auto.arima()} function compares either AIC, AICc or BIC value and suggests best ARIMA or SARIMA models for a given time series data. Analogously, the \code{ets()} function considers the best exponential smoothing state space model automatically and predicts future values.


\begin{table}
\centering
\begin{tabular}{c|c|c|c}
\toprule
Functions&\code{psf()}&\code{auto.arima()}&\code{ets()}\\

\midrule
RMSE&2.077547&2.340092&32.78256\\

\bottomrule
\end{tabular}
\caption{Comparison of time series prediction methods with respect to RMSE values for \code{nottem} dataset.}
\label{tab:label1}
\end{table}

\begin{table}
\centering
\begin{tabular}{c|c|c|c}
\toprule
Functions&\code{psf()}&\code{auto.arima()}&\code{ets()}\\

\midrule
RMSE&22.11279&41.14366&52.29985\\

\bottomrule
\end{tabular}
\caption{Comparison of time series prediction methods with respect to RMSE values for \code{sunspots} dataset.}
\label{tab:label2}
\end{table}

Tables \ref{tab:label1} and \ref{tab:label2} show comparisons for \code{psf()}, \code{auto.arima()} and \code{ets()} functions when using the Root Mean Square Error (RMSE) parameter as metric, for \code{nottem} and \code{sunspots} datasets, respectively. In order to avail more accurate and robust comparison results, error values are calculated for 10 times and the mean value of error values for methods under comparison are shown in Tables \ref{tab:label1} and \ref{tab:label2}. These tables clearly state that \code{psf()} function is able to outperform the comparative time series prediction methods.

Additionally, the reader might want to refer to the results published in the original work \cite{martinez2011energy}, in which it was shown that PSF outperformed many different methods when applied to electricity prices and demand forecasting.

\section{5. Conclusions}\label{sec:Conclusions}

This article is a thorough description of the PSF R package. This package is introduced to promote the algorithm Pattern Sequence based Forecasting (PSF) which was proposed by \citet{martinez2008lbf} and then modified and improved by \citet{martinez2011energy}. The functions involved in the PSF package can be tested with time series data in any format like vector, time series, list, matrix or data frame. The aim of the PSF package is to simplify the calculations and to automate the steps involved in prediction while using PSF algorithm. Illustrative examples and comparisons to other methods are also provided.

\section{Acknowledgements}

We would like to thank our colleague, Mr. Sudhir K. Mishra, for his valuable comments and suggestions to improve this article. Thanks also goes to CRAN maintainers for needful comments on the submitted package. Finally, this study has been partially funded by the Spanish Ministry of Economy and Competitiveness and by the Junta de Andaluc\'ia under projects TIN2014-55894-C2-R and P12-TIC-1728, respectively.

\bibliography{bokde}

\address{Neeraj Bokde\\
  Visvesvaraya National Institute of Technology, Nagpur\\
  North Ambazari Road, Nagpur\\
  India\\}
\email{neeraj.bokde@students.vnit.ac.in}

\address{Gualberto Asencio-Cort\'es\\
  Universidad Pablo de Olavide\\
  ES-41013, Sevilla\\
  Spain\\}
\email{guaasecor@upo.es}

\address{Francisco Mart\'inez-\'Alvarez\\
  Universidad Pablo de Olavide\\
  ES-41013, Sevilla\\
  Spain\\}
\email{fmaralv@upo.es}

\address{Kishore Kulat\\
  Visvesvaraya National Institute of Technology, Nagpur\\
  North Ambazari Road, Nagpur\\
  India\\}
\email{kdkulat@ece.vnit.ac.in}

\end{article}

\end{document}